\documentclass[letterpaper, 10 pt, conference]{ieeeconf}  

\IEEEoverridecommandlockouts                              

\usepackage{times}
\usepackage{epsfig}
\usepackage{graphicx}
\usepackage{amsmath}
\usepackage{amssymb}
\usepackage[inkscapearea=page]{svg}
\usepackage{algorithm}
\usepackage{algpseudocode}
\usepackage{import}
\usepackage{subcaption}

\newcommand{\etal}{\textit{et al.}}

\usepackage{tikz}

\usepackage{academicons}
\definecolor{orcidlogocol}{HTML}{A6CE39}
\definecolor{lime}{HTML}{A6CE39}

\DeclareRobustCommand{\orcidicon}{%
    \begin{tikzpicture}
    \draw[lime, fill=lime] (0,0) 
    circle [radius=0.16] 
    node[white] {{\fontfamily{qag}\selectfont \tiny ID}};
    \draw[white, fill=white] (-0.0625,0.095) 
    circle [radius=0.007];
    \end{tikzpicture}
    \hspace{-2mm}
}
\newcommand{\orcidWalter}{\href{https://orcid.org/0000-0003-4565-1272}{\orcidicon}}
\newcommand{\orcidMarcus}{\href{https://orcid.org/0000-0001-8984-2551}{\orcidicon}}
\newcommand{\orcidKnoll
}{\href{https://orcid.org/0000-0003-4840-076X}{\orcidicon}}

\usepackage{adjustbox}

\usepackage[pagebackref=true,breaklinks=true,letterpaper=true,colorlinks,bookmarks=false]{hyperref}

\usepackage[capitalize]{cleveref}

\crefname{section}{Sec.}{Secs.}
\Crefname{section}{Section}{Sections}

\crefname{table}{Tab.}{Tabs.}
\Crefname{table}{Table}{Tables}

\crefname{equation}{Eq.}{Eqs.}
\Crefname{equation}{Equation}{Equations}

\makeatletter
\newcommand*{\emails}[2][@tum.de]{%
    \def\@tempa{\@gobble}%
    \@for\qrr@email:=#2\do{%
        \edef\@tempb{\noexpand\href{mailto:\qrr@email #1}{\qrr@email}}%
        \edef\@tempa{\unexpanded\expandafter{\@tempa}{, }\unexpanded\expandafter{\@tempb}}}%
    \{\@tempa\}#1%
}    

\title{\LARGE \bf
Real-Time and Robust 3D Object Detection Within Roadside LiDARs Using Domain Adaptation
}

\author{
Walter Zimmer$^{1}$\orcidWalter, Marcus Grabler$^{1,2}$\orcidMarcus and Alois Knoll$^{1}$\orcidKnoll
\thanks{*This research was supported by the Federal Ministry of Education and Research in Germany within the project \textit{AUTOtech.agil}, Grant Number: 01IS22088U.}
\thanks{$^{1}$The authors are with the Informatics Faculty, Technical University of Munich (TUM), 85748 Garching-Hochbrueck, Germany
        {\tt\small \href{mailto:walter.zimmer@cs.tum.edu}{walter.zimmer@cs.tum.edu}, 
        \emails{grabler,knoll}
        }}%
\thanks{$^{2}$Autonomous Reply, Riesstra{\ss}e 22, 80992 Munich, Germany
        {\tt\small \href{mailto:m.grabler@reply.de}{m.grabler@reply.de}}}%
}

\begin{document}

\maketitle
\thispagestyle{empty}
\pagestyle{empty}

\begin{abstract}
   This work aims to address the challenges in domain adaptation of 3D object detection using roadside LiDARs. We design \textit{DASE-ProPillars}, a model that can detect objects in roadside LiDARs in real-time. Our model uses \textit{PointPillars} as the baseline model with additional modules to improve the 3D detection performance. To prove the effectiveness of our proposed modules in \textit{DASE-ProPillars}, we train and evaluate the model on two datasets, the open source A9 dataset and a semi-synthetic roadside A11 dataset created within the \textit{Regensburg Next} project. We do several sets of experiments for each module in the \textit{DASE-ProPillars} detector that show that our model outperforms the \textit{SE-ProPillars} baseline on the real A9 test set and a semi-synthetic A9 test set, while maintaining an inference speed of 45 Hz (22 ms) that allows to detect objects in real-time. We apply domain adaptation from the semi-synthetic A9 dataset to the semi-synthetic A11 dataset from the \textit{Regensburg Next} project by applying transfer learning and achieve a $3D~mAP@0.25$ of 93.49\% on the Car class of the target test set using 40 recall positions. 
\end{abstract}


\section{Introduction}
High quality and balanced data is crucial to achieve high accuracy in deep learning applications. The creation of labeled data of roadside LiDARs is a difficult task. Considering the high labor cost of manually labeling 3D LiDAR point clouds, we need to find a solution to deal with small datasets. Publicly available LiDAR datasets were recorded and labeled from a vehicle perspective which makes is difficult to apply these trained detectors on roadside LiDARs. The focus of this work lies in the area of domain adaptation to tackle the domain shift problem. How can a neural network that was trained in one operational design domain (ODD), e.g. an urban area like in the A9 dataset \cite{cress2022a9}, be adapted to a slightly different domain, e.g. an intersection in a different city with different LiDAR sensors and mounting positions? This process is known as transfer learning -- training a model on a large dataset (source domain) and fine-tuning it on another dataset (target domain). Another challenge is real-time 3D object detection on roadside LiDARs, i.e. to detect objects at a high frame rate to prevent accidents. This highly depends on the LiDAR type, the rotation rate, and the number of 3D points. The final challenge this work is dealing with is a robust 3D detection of all traffic participants. Detecting small and occluded objects at different weather conditions and rare traffic scenarios is a highly important research area to increase safety of automated vehicles.\\
\begin{figure}[t]
    \centering
    \includegraphics[width=1.0\linewidth]{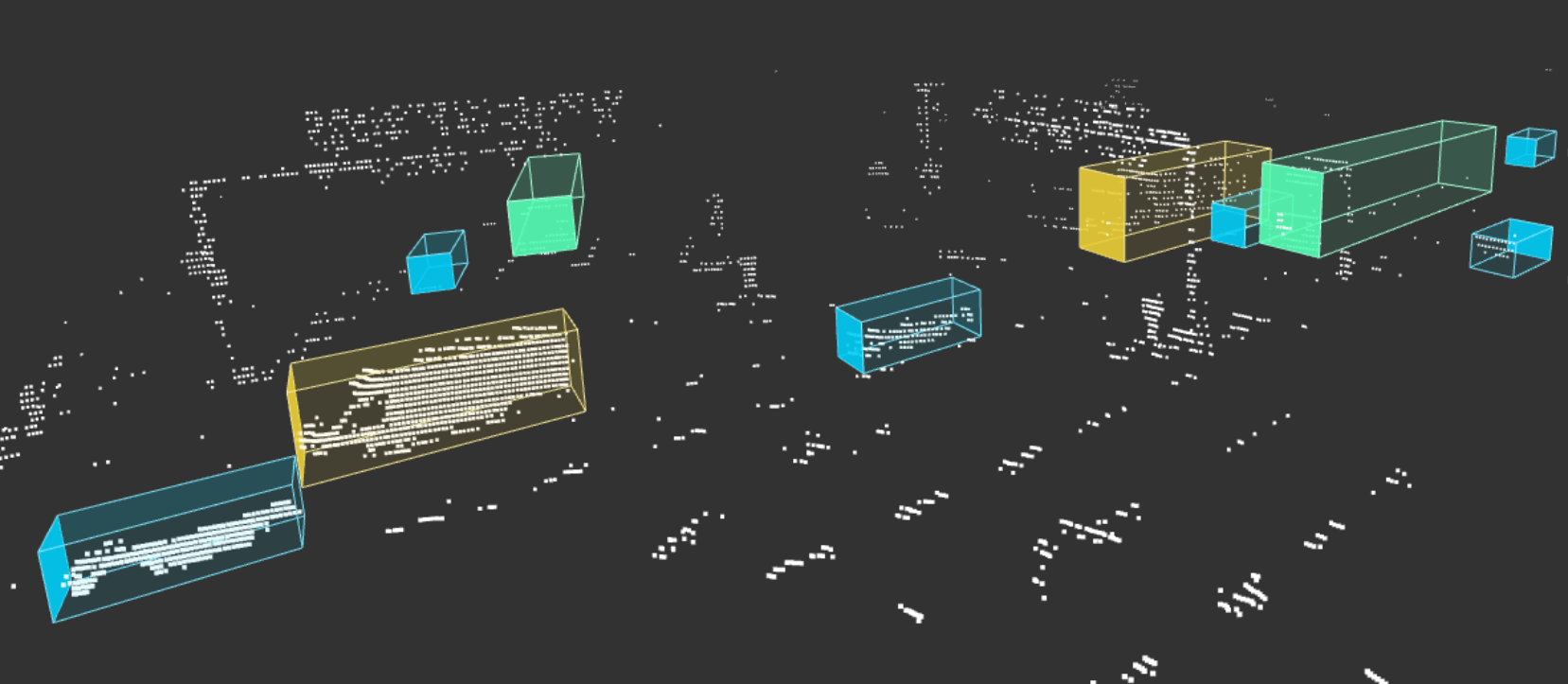}
    \caption{Labeled 3D objects in the semi-synthetic \textit{proSynthSemi} dataset.}
    \label{fig:prosynthsemi_objects}
\end{figure}

\begin{figure*}
    \centering
    \includegraphics[clip, trim=0cm 2.2cm 0cm 0cm,width=\textwidth]{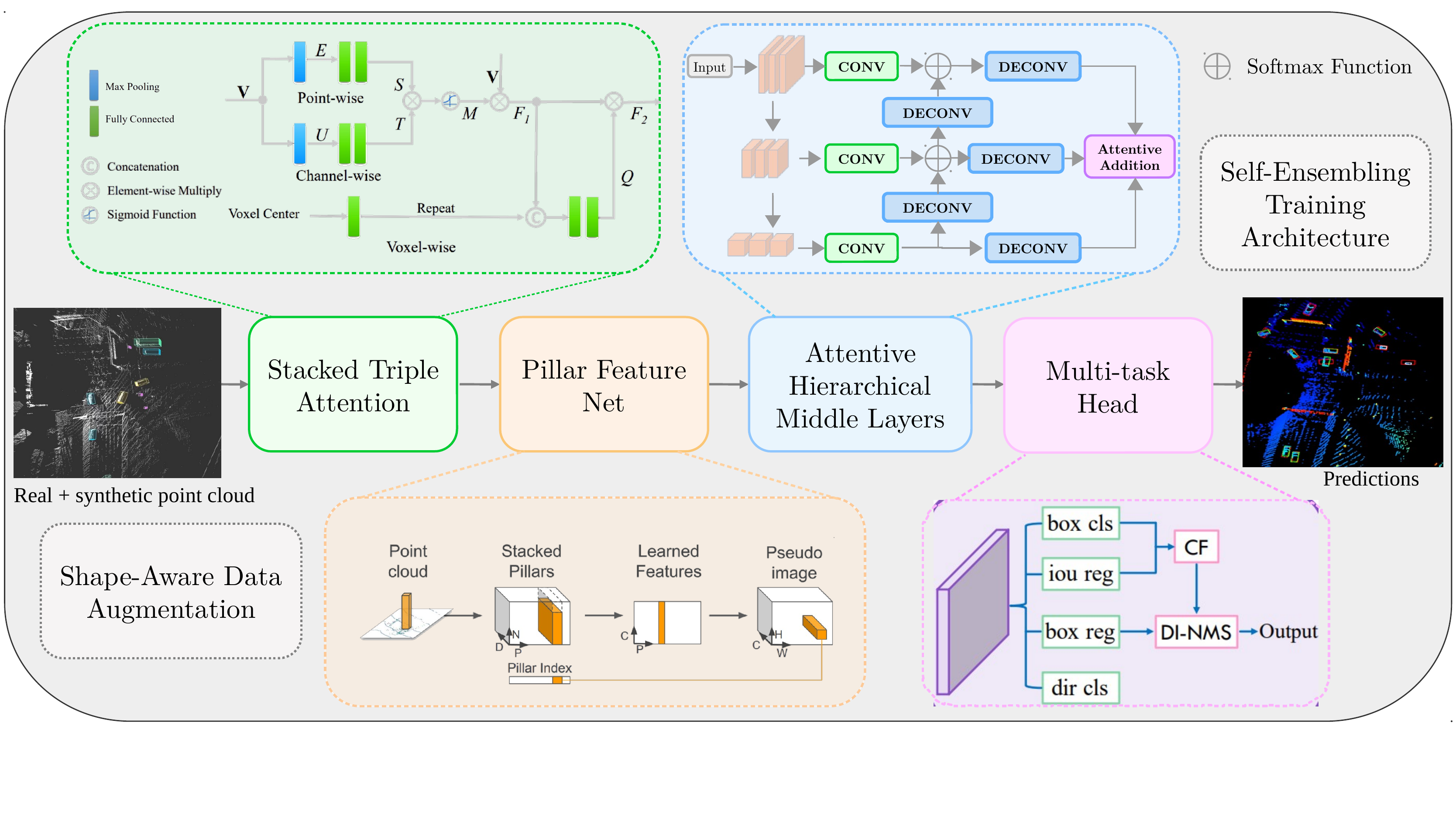}
    \caption{Overview architecture of \textit{DASE-ProPillars}, a LiDAR-only single-stage pillar-based 3D object detector. The detector is based on \textit{PointPillars} \cite{lang2019pointpillars}, with the following five extensions. 1) Data Augmentation (Shape-Aware \cite{zheng2021se}, Dropout, upsampling and Gaussian Noise). 2) Stacked Triple Attention Mechanism \cite{liu2020tanet}. 3) Attentive Hierarchical Middle Layers. 4) Multi-task detection head. 5) Self-Ensembling Training Architecture \cite{zheng2021se}. The stacked triple attention module extracts features from the semi-synthetic point cloud using the triple attention mechanism, including channel-wise, point-wise, and voxel-vise attention to enhance the learned features. The pillar feature net turns point-wise features into pillar features and scatters the pillar features into a pseudo image. The hierarchical middle layers perform 2D convolution operations on the pseudo image. Hierarchical feature maps are concatenated with attentive addition. Finally, the multi-task head is used for the final prediction, that includes an IoU prediction to alleviate the misalignment between the localization accuracy and classification confidence. In addition, we introduce two new training techniques: the shape-aware data augmentation module and the self-ensembling teacher and student training framework. 
    }
    \label{fig:architecture}
\end{figure*}
We create and open source a large semi-synthetic roadside dataset with 7,000 labeled point cloud frames (see \cref{fig:prosynthsemi_objects}). This dataset is balanced in terms of object classes and contains a high variety so that objects can be detected in different scenarios and different environment conditions. We analyze whether transfer learning from a larger roadside LiDAR dataset, such as the A9 dataset, can improve the model performance on other roadside datasets. The first batch of the published A9 dataset includes 459 manually labeled point cloud frames and contains 3,104 labeled 3D objects. In this work we propose a single-stage 3D object detector, train it in one domain and finetune it on a different domain. An intersection, that is part of the A9 Test Stretch for Autonomous Driving \cite{krammer2022providentia,lakshminarasimhan2020c,cress2021intelligent}, is equipped with five LiDAR sensors (see \cref{fig:intersection}), in order to represent a real-time digital twin of the traffic.
This work provides a domain adaptation solution for the single LiDAR detection task. The main contributions in this work are summarized as follows:
\begin{itemize}
    \item We propose a robust single-stage LiDAR-only detector based on \textit{PointPillars}. We introduce five extensions to improve PointPillars and evaluate the performance of the model on the test set of the A9 \cite{cress2022a9}, the A11 and D16 dataset from the \textit{Regensburg Next} project \cite{torunskyPilotprojektMitVorbildcharakter}. 
    \item We introduce a synthetic data generation module for the CARLA simulator \cite{dosovitskiy2017carla}, that converts the labels to the OpenLABEL format \cite{OpenLABELStandard}.
    \item We propose a novel domain adaptation technique, the semi-synthetic data generation method, which decreases the sim-to-real gap, as shown in experiments.
    \item We create a semi-synthetic dataset, called \textit{proSynthSemi}, with 7,000 labeled LiDAR point cloud frames using the CARLA simulator \cite{dosovitskiy2017carla} and train our model on that. In addition, we provide two synthetic datasets, A11 and D16, with 2,581 and 3,315 labeled frames respectively. 
    \item Experiments show that our \textit{DASE-ProPillars} model outperforms the \textit{SE-ProPillars} \cite{sepropillars} model by 30.56\% 3D mAP on the A9 test set (Car class), while the inference speed is maintained at 45 Hz (22 ms). 
\end{itemize}

\section{Related work}

First, we compare one-stage and two stage methods, types of point cloud representations, single- and multi-frame approaches, supervised and unsupervised as well as center and anchor-based methods. In the second part we analyze the importance of data augmentation and the generation of synthetic data to solve the domain shift problem.
\subsection{3D Object Detection Models}
According to the form of feature representation, LiDAR-only 3D object detectors can be divided into four main streams, i.e. point-based, voxel-based, range-view based and multi-view-based methods \cite{walter2022a}. In point-based methods, features maintain the form of point-wise features, either by a sampled subset or derived virtual points. 
PointRCNN \cite{pointrcnn} uses a PointNet++ backbone \cite{pointnet++} to extract point-wise features from the raw point cloud, and performs foreground segmentation. For each foreground point, it generates a 3D proposal followed by a point cloud ROI pooling and a canonical transformation-based bounding box refinement process. Point-based methods usually have to deal with a huge amount of point-wise features, which leads to a lower inference speed. To accelerate point-based methods, 3DSSD \cite{3dssd} introduces feature farthest-point-sampling (F-FPS), which computes the feature distance for sampling, instead of Euclidean distance in traditional distance farthest-point-sampling (D-FPS). The inference speed of 3DSSD is competitive with voxel-based methods.

\begin{figure*}[t]
    \centering
    \includegraphics[height=3.2cm]{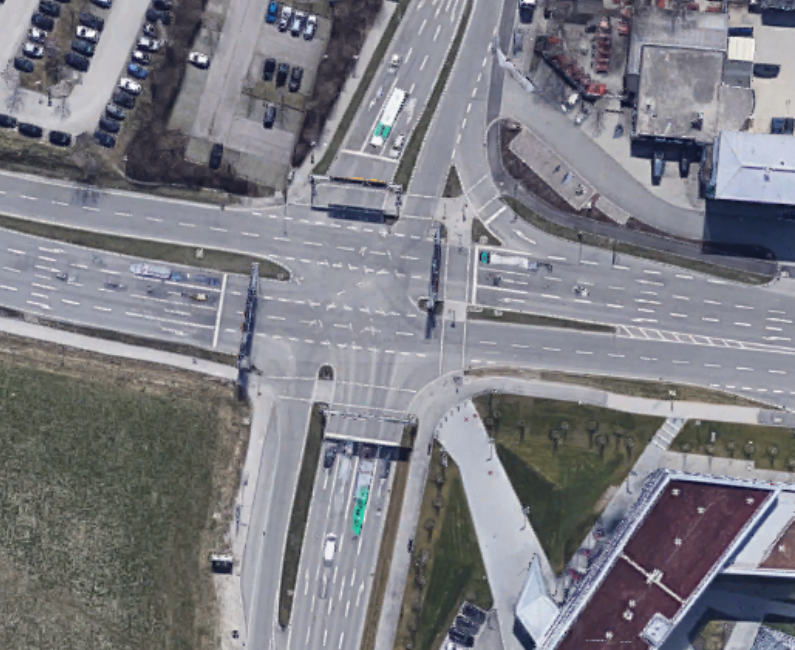}
    \includegraphics[height=3.2cm]{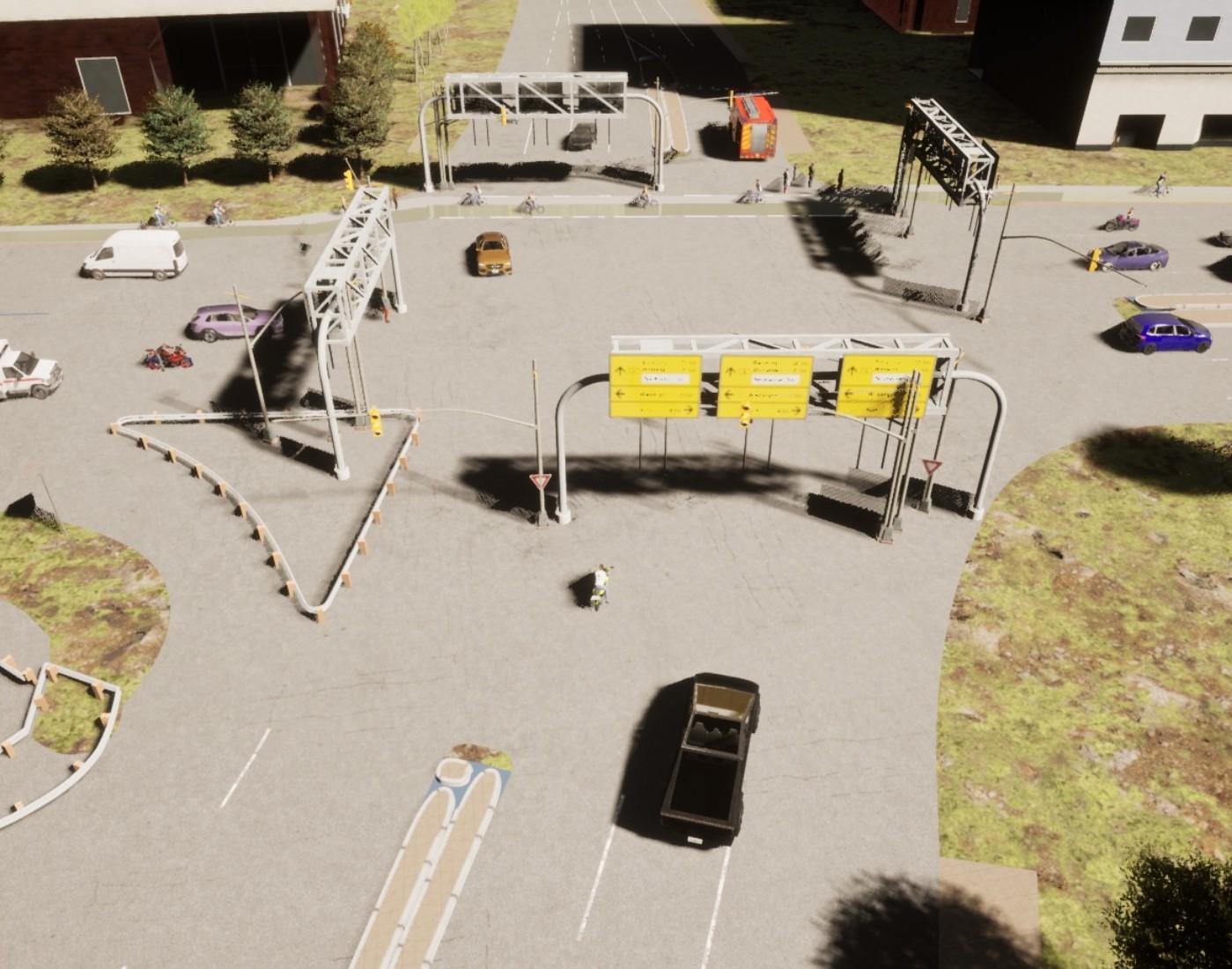}
    \includegraphics[height=3.2cm]{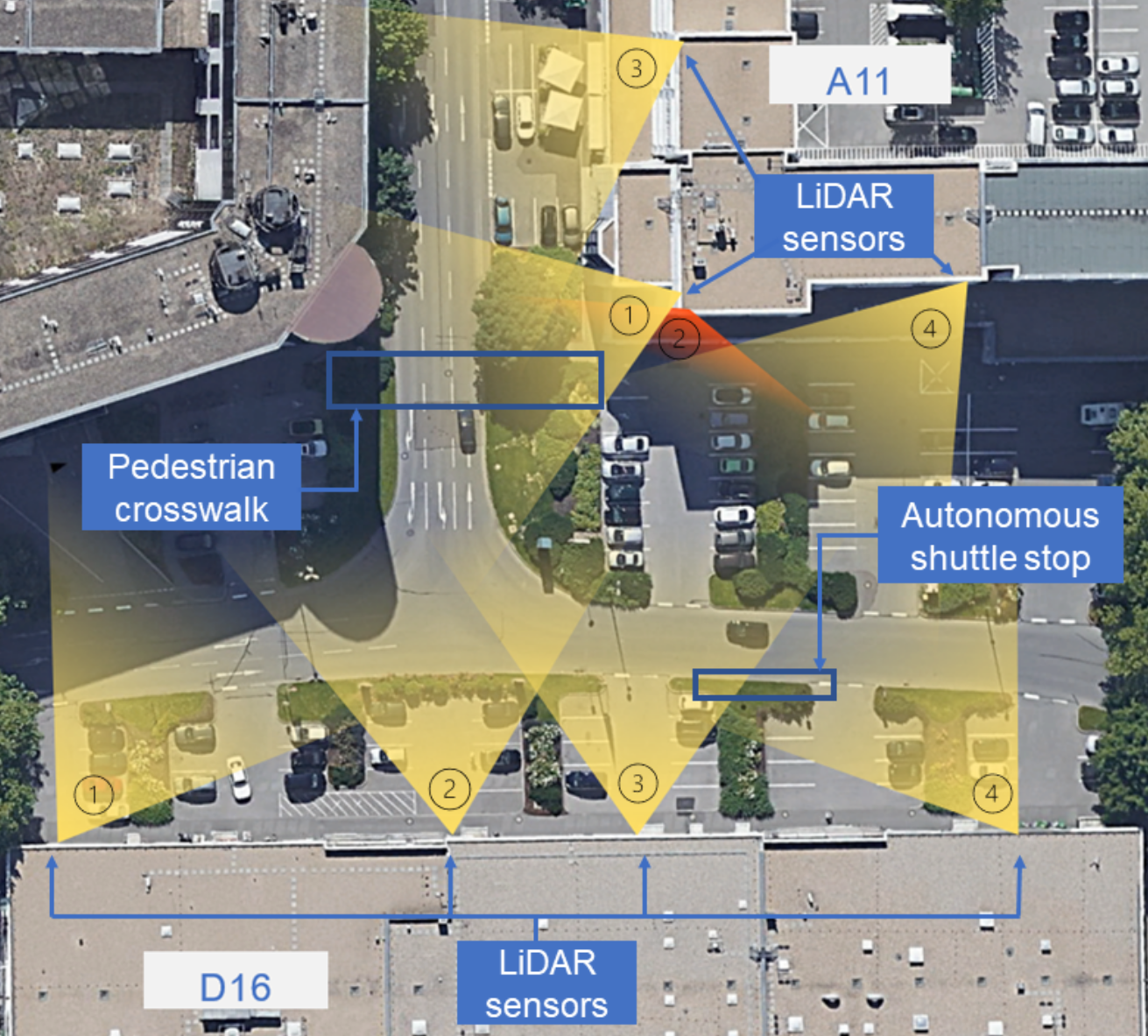}
    \includegraphics[height=3.2cm]{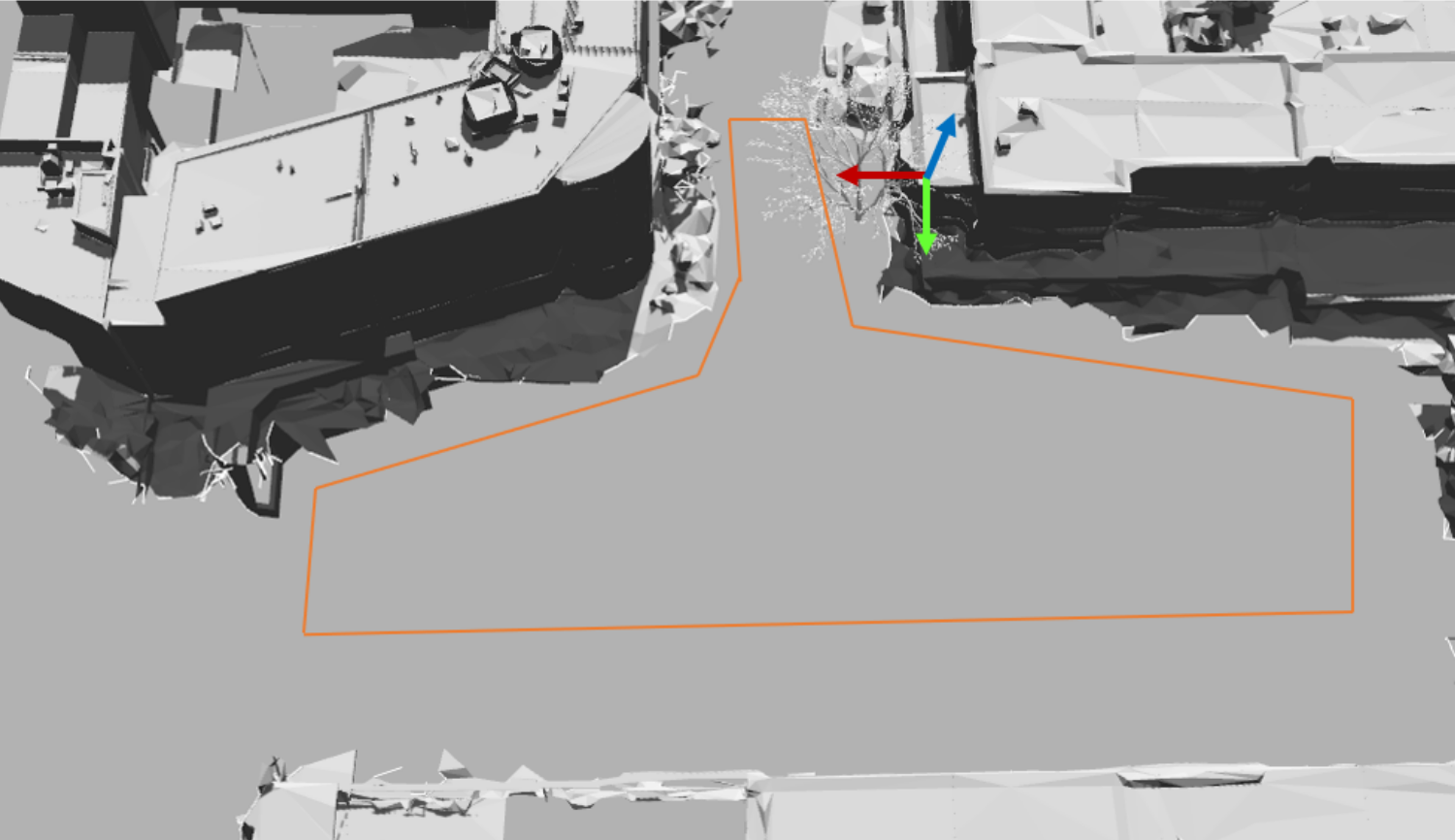}
    \caption{a) Birds-eye view (BEV) of the S110 intersection that is part of the A9 test stretch. b) S110 intersection modeled in the CARLA simulator and Unreal Engine. c) Intersection in Regensburg that was used to create the A11 and D16 synthetic datasets. d) Reconstruction of the intersection in Regensburg.}
    \label{fig:intersection}
\end{figure*}

SECOND \cite{second} proposes a sparse convolutional middle extractor \cite{subsparseCNN} to speed up inference time. In PointPillars \cite{lang2019pointpillars}, the point cloud is divided into pillars (vertical columns), which are special voxels without partition along the z-direction. The feature map of pillars is a pseudo-image so that 2D convolutions can be used. PointPillars runs with 62 FPS using TensorRT. 

SA-SSD \cite{sassd} adds a detachable auxiliary network to the sparse convolutional middle layers to predict a point-wise foreground segmentation and a center estimation task to provide a point-level supervision. It also proposes a part-sensitive warping (PS-Warp) operation as an extra detection head. It can alleviate the misalignment between predicted boxes and classification confidence maps, since they are generated by two different convolutional layers in the detection head. 

CIA-SSD \cite{zheng2020cia} designs an IoU-aware confidence rectification module, using an additional convolutional layer in the detection head to make IoU predictions. The predicted IoU value rectifies the classification score. By introducing only one additional convolutional layer, it is more lightweight than SA-SSD. 

SE-SSD \cite{zheng2021se} proposes a self-ensembling one-stage post-training framework, where a pre-trained teacher model produces predictions that serve as soft targets in addition to the hard targets from the label. These predictions are matched with student's predictions by their IoU and supervised by the consistency loss. Soft targets are closer to the predictions from the student model and therefore help the student model to finetune its predictions. The Orientation-Aware Distance-IoU Loss (OD-IoU) is proposed to replace the traditional smooth-$L_1$ loss of box regression in the post training, in order to provide a fresh supervisory signal. This OD-IoU loss emphasizes the orientation of the bounding boxes as well as the alignment of the center points. SE-SSD also designs a shape-aware data augmentation module to improve the generalization ability of the student model. This module performs dropout, swapping and sparsification of points. This data augmentation is applied to the point cloud data the student model is trained on. In this way, by using both a teacher and a student single-stage object detector, the framework can boost the precision of the detector significantly without incurring extra computation during the inference.\\
Pyramid R-CNN \cite{mao2021pyramid} concentrates on handling the sparsity and non-uniform distribution of point clouds. The authors propose a novel pyramid RoI-head second-stage module, that extracts the features from sparse points of interest. Utilizing features only inside the RoIs performs well in 2D detection models mainly for two reasons. First, the input feature map is dense and second, the collected pixels have large receptive fields. However, in 3D models, the points of interest are sparse and non-uniformly distributed inside the RoIs. Therefore, accurately inferring the sizes and categories of objects becomes hard when collecting features of few individual points and not gathering enough information from neighbours. Pyramid RoI effectively solves this problem by constructing a pyramid grid structure that contains the RoI-grid points both inside and outside RoIs, so that both fine-grained shape structures for accurate box refinement as well as large context information for identifying incomplete objects can be captured by grid points inside RoIs and outside RoIs, respectively. The authors performed experiments on the KITTI Dataset and the Waymo Open Dataset \cite{sun2020scalability}. They showed that Pyramid R-CNN outperforms other 3D detection models on these two datasets. On the Waymo Open Dataset their Pyramid-PV model achieves 81.77\% L1 mAP.\\
As conventional 3D convolutional backbones in voxel-based 3D detectors are not able to efficiently capture extensive context information, Voxel Transfomer (VoTr) \cite{VoTr} is proposed to resolve this issue by introducing a voxel-based transformer backbone for 3D object detection from point clouds. It consists of a series of sparse voxel modules, which extract features at empty locations and thus are responsible for the downsampling of the voxel-grids and submanifold voxel modules, which perform multi-head self attention strictly on non-empty voxels to keep the original 3D structure with increasing the receptive fields. The attention mechanism is split up into two components, called local attention and dilated attention. Local attention focuses on the neighboring region to preserve detailed information. Dilated attention obtains a large attention range with only a few attending voxels by gradually increasing the search step. VoTr can be applied to single-stage and two-stage detectors. Comparing with the backbone of the respective module, the AP results on the KITTI test set, calculated by 40 recall positions for the car class, are slightly increased.\\
Signal miss, external and self occlusion often cause shape misses for disordered point clouds. Behind the Curtain Detector (BtcDet) \cite{BtcDet} deals with this problem by learning the object shape priors and estimate the complete object shapes including the partially occluded points. Therefore, points are filled into the labeled bounding boxes and using the recovered shape miss, the detection results are improved. For the training process, the complete object shapes are approximated using the ground truth labels of the corresponding objects. For cars and cyclist, the objects points are mirrored against the middle section plane of the bounding box and a heuristic H(A,B) determines if a source object B covers most parts of a target object A. The heuristic also provides points that can fill the target object's shape miss. The detection pipeline of BtcDet is built up as followed: First, the regions of occlusion and signal miss have to be identified after the spherical voxelization for the point cloud. Then, a shape occupancy network estimates the probability of object shape occupancy using the created training targets, which consists of the approximated complete object shapes. The extracted point cloud 3D features are then sent to a region proposal network to generate 3D proposals, which are refined in the last step, the proposal refinement step.

\subsection{Domain Adaptation}
Domain adaptation is a type of transfer learning and aims to transfer knowledge from a source domain, for which annotated data is available to a target domain, for which no or only less annotated data is available. Semi-supervised domain adaptation uses a few labeled examples from the target domain to learn a target model and unsupervised domain adaptation exploits only the labeled data from the source domain without having any annotated target domain data \cite{TransferLearningFromSynthToReal}. The domain adaptation methods can be divided into four different approaches, which are either data-driven, such as domain-invariant data representation, domain mapping, and normalization statistics or model-driven like the domain-invariant feature learning \cite{surveyDA}. With the use of these domain adaptation methods, the gap between the source and the target domain should be mitigated \cite{domainandmodalitygaps2021}. 

Wang \etal \cite{traininger} propose a dataset-to-dataset, semi-supervised domain adaptation for 3D object detection and provide a baseline for 3D object detection adaptation across countries using normalization statistics domain adaptation methods. As only annotated datasets are considered, few-shot fine-tuning enables to increase the accuracy by selecting 10 labeled scenes from the target domain, which are added to the source domain during training. As the car sizes vary in several countries, the object sizes of the source domain differ from the object sizes of the target domain. Therefore, the already trained object detector is modified that the predicted box sizes can better match the previously determined target statistics. To fit the adjusted box size, the corresponding labels are scaled up or shrunk down. New point clouds with the associated labels whose sizes are similar to the target domain data are generated. This step is called statistical normalization. 

ST3D \cite{ST3D} provides a self-training pipeline for unsupervised domain adaptation on 3D object detection from point clouds, where no annotated data in the target domain is available. As the sizes of the objects vary in different datasets due to the geographical location in which the data were recorded, ST3D proposes pre-training of the 3D detector on source domain with random object scaling (ROS) strategy to mitigate the negative effects of source domain bias. Using the 3D object detector, pseudo labels for the unlabeled target data are generated. The quality-aware triplet memory bank (QTMB) modules parses the object predictions to the pseudo labels. However, the negative impacts on pseudo labeled objects lead to noisy supervisory information and instability for self-training. The memory bank updates the pseudo labels that also serve as labels for subsequent model training. The curriculum data augmentation (CDA) module allows to generate gradually increasingly diverse and potentially hard examples to improve the model. This enables the model to learn from challenging samples while making the examples more difficult during training.

\section{Approach}
We design a real-time LiDAR 3D object detector (\textit{DASE-ProPillars}) to solve the domain shift problem. The architecture of our \textit{DASE-ProPillars} model is shown in Fig. \ref{fig:architecture}.

\textbf{Data Generation.}
 We created a semi-synthetic dataset (\textit{proSynthSemi}) with 7,000 point cloud frames using the CARLA simulator and train our \textit{DASE-ProPillars} model on it. Fig. \ref{fig:intersection} shows the intersection with generated traffic in the CARLA simulator. A simulated LiDAR sensor represents a real Ouster OS1-64 (gen. 2) LiDAR sensor with 64 channels and a range of 120 m. In the simulation, we add Gaussian noise (see \cref{eq:gaussian_noise}) $N(\mu,\sigma^2)$ with mean $\mu=0$ and standard deviation $\sigma=0.1$ to disturb each point along the vector of its raycast. 
 \begin{equation}\label{eq:gaussian_noise}
    p(z) = \frac{1}{\sigma\sqrt{2\pi}}e^{\frac{(z-\mu)^2}{2\sigma^2}}
 \end{equation}
 The LiDAR emits 1.31M points per second and runs at 10 Hz (131,000 points per frame). We store the extracted point clouds in .pcd files and labels in .json files according to the OpenLABEL standard \cite{OpenLABELStandard}. To get more realistic point clouds, the points of the objects in the simulated point clouds are extracted and included into a background of a point cloud captured by a real Ouster OS1-64 (gen.2) LiDAR, which does not contain any objects. Before the simulated object points are inserted into the real background point set, those points in the real point cloud are cut out, which are inside the simulated objects or below, respectively the points of the ground plane. As the height profiles of simulated and real data do not coincide, the z-coordinate of the objects must be adjusted in order to place the objects on the ground plane. We use the \textit{RANSAC} algorithm \cite{ransac} to determine the ground plane of the real point cloud and thus, calculating the corresponding height profile. Subsequently, a pipeline is applied to the background points using both Gaussian noise and drop out of points in order to get more variance in the point clouds. This gives the advantage to get more realistic point clouds and having labeled objects from the simulation. 

\textbf{Normalization.} As the bounding boxes of the classes differ between the semi-synthetic and real A9 dataset, the boxes are normalized to an average size of the bounding boxes for each class.  Since manually labeled data can often result in incorrect sizes of the length, width and height of the bounding box, normalized bounding boxes of the synthetic labels are used instead. For synthetic data, the exact dimensions can be extracted directly from the simulation. With normalization, the distinction between the individual classes can be improved, which is useful, e.g. for the \textit{Van} and \textit{Car} classes. In addition, the normalized sizes are used in a domain adaptation to adjust the source domain data to the target domain data.

\textbf{Voxelization.} We divide the raw point cloud into vertical pillars before feeding them into a neural network. These are special voxels that are not split along the vertical axis. Pillars have these advantages over voxels: A pillar-based backbone is faster than a voxel-based backbone due to fewer grid cells. Time consuming 3D convolutional middle layers are also being eliminated and instead 2D convolutions are being used. We also do not need to manually tune the bin size along the z-direction hyperparameter. If a pillar contains more points than specified in the threshold, then the points are being subsampled to the threshold using farthest point sampling \cite{eldar1997farthest}. If a pillar contains fewer points than the threshold, then it is padded with zeros to make the dimensions consistent. Due to the sparsity issue most of the pillars are empty. We record the coordinates of non-empty pillars according to the pillar's center index. Empty pillars are not being considered during the feature extraction until all pillars are being scattered back to a pseudo image for 2D convolution. For experiments, we set the voxel size to (0.2, 0.2, 6.0) m, for which the height (6.0 m) must correspond to the detection range in the z-axis. The maximum number of points per voxel is set to 40 and if a voxel contains more, the points are subsampled using the farthest sampling method. We also limit the number of voxels to 20,000.

\textbf{Stacked Triple Attention.} The \textit{Stacked Triple Attention} module is used for a more robust and discriminative feature representation. Originally introduced in \textit{TANet} \cite{liu2020tanet} by Liu \etal, the stacked triple attention module enhances the learning of hard to detected objects and deals better with noisy points. This method can be applied on both, voxel and pillar-based point clouds. The attention mechanism in this module follows the Squeeze-and-Excitation pattern \cite{hu2018squeeze}. If channel-wise attention is applied to an input tensor with shape ($H \times W \times C$), then first a global pooling operation (max pooling) is used to pool the tensor to shape ($1 \times 1 \times C$), called squeeze operation. Then, two fully connected (FC) layers are applied to the squeezed tensor attention score, called excitation operation. Between the two FC layers, the feature dimension is reduced and then recovered with a reduction ratio which forms a bottleneck structure. After that, a sigmoid function is applied to get the attention score. Finally, the ($1 \times 1 \times C$) tensor is multiplied element-wise to get the original ($H \times W \times C$) feature.

The input to the module is a $(P \times N \times C)$ tensor, where $P$ is the number of non-empty pillars, $N$ is the maximum number of points in each pillar, and $C$ is the dimension of the input point-wise feature. At the beginning, we have a 9-dimensional ($C = 9$) feature vector $(x, y, z, r, x_c, y_c, z_c, x_p, y_p)$, where $x, y, z$ are the coordinates of the point, r is the intensity, $x_c, y_c, z_c$ are the distance to the arithmetic mean of all points inside the pillar, $x_p, y_p$ are the location of the pillar from the pillars center.
The triple attention (TA) module extracts features inside each pillar, using point-wise, channel-wise and voxel-wise attention. All three attention scores are combined to form the final output feature. 
To further exploit the multi-level feature attention, two triple attention modules are stacked with a structure similar to the skip connections in ResNet \cite{he2016deep}. The first module takes the raw point cloud 9-dim features as input, while the second one works on the extracted high dimensional features. For each TA module the input is concatenated or summed to the output to fuse more feature information. Each TA module is followed by a fully connected layer to increase the feature dimension. Inside the TA modules, the attention mechanism only re-weights the features, but does not increase their dimensions.

\textbf{Pillar Feature Net.} We choose \textit{PointPillars} \cite{lang2019pointpillars} as our baseline and improve its 3D detection performance at the expense of inference time. \textit{PointPillars} runs at 42 Hz without the acceleration of \textit{TensorRT}. Since there is a trade-off between speed and accuracy, we can further boost the accuracy by incorporating additional modules without sacrificing the inference speed too much. The pillar feature net (PFN) shown in Fig. \ref{fig:architecture} takes pillars as input, extracts pillar features, and scatters pillars back to a pseudo image for 2D convolution operations in the middle layers. The pillar feature net acts as an additional feature extractor to the stacked triple attention module. The point-wise pillar-organized features from the stacked TA module with shape ($P \times N \times C$) are fed to a set of PFN layers. Each PFN layer is a simplified PointNet \cite{qi2017pointnet}, which consists of a linear layer, Batch-Norm \cite{ioffe2015batch}, ReLU \cite{nair2010rectified}, and max pooling. The max-pooled features are concatenated back to the ReLU's output to keep the point-wise feature dimension inside each pillar, until the last FPN layer. The last FPN layer makes the final max pooling and outputs a ($P \times C$) feature as the pillar feature. Pillar features are then scattered back to the original pillar location, forming a ($C \times H \times W$) pseudo image, where $H$ and $W$ are the height and width of the pillar grid. Here the location of empty pillars is padded with zeros.

\textbf{Attentive Hierarchical Middle Layers.} We exchange the default backbone of \textit{PointPillars} with an \textit{Attentive Hierarchical Backbone} to perform 2D convolution on the pseudo image from the pillar feature net. In the first stage, the spatial resolution of the pseudo image is gradually downsampled by three groups of convolutions. Each group contains three convolutional layers, where the first one has a stride of two for downsampling, and the two subsequent layers act only for feature extraction. After downsampling, deconvolution operations are applied to recover the spatial resolution. Deconvolutional layers (marked with an asterix) recover the size of feature maps with stride 2 and element-wise add them to upper branches. The remaining three deconvolutional layers make all three branches have the same size (half of the original feature map). Then the final three feature maps are combined by an attentive addition to fuse both, spatial and semantic features. The attentive addition uses the plain attention mechanism. All three feature maps are being passed through a convolutional operation and are channel-wise concatenated as attention scores. The softmax function generates the attention distribution and feature maps are multiplied with the corresponding distribution weight. The element-wise addition in the end gives the final attention output, a ($C \times H/2 \times W/2$) feature map.

\textbf{Multi-task Head.}
The multi-task head outputs the final class (based on a confidence score), the 3D box position ($x, y, z$), dimensions ($l, w, h$), rotation ($\theta$) and the direction of the detected object. The direction (front/back) is being classified to solve the problem that the sine-error loss \cite{yan2018second} cannot distinguish flipped boxes. Four convolutional layers operate on the feature map separately. One of the four heads is the IoU prediction head that predicts an IoU between the ground truth bounding box and the predicted box. It was introduced in CIA-SSD \cite{zheng2020cia} to deal with the misalignment between the predicted bounding boxes and corresponding classification confidence maps. The misalignment is mainly because these two predictions are from different convolutional layers. Based on this IoU prediction, we use the confidence function (CF) to correct the confidence map and use the distance-variant IoU-weighted NMS (DI-NMS) module post-process the predicted bounding boxes. The distance-variant IoU-weighted NMS is designed to deal with long-distance predictions, to better align far bounding boxes with ground truths, and to reduce false-positive predictions. If the predicted box is close to the origin of perspective, we give higher weights to those box predictions with high IoU. If the predicted box is far, we give relatively uniform weights, to get a more smooth final box.

\textbf{Data Augmentation.}
Data augmentation has proven to be an efficient way to better exploit the training dataset and help the model to be more generalized. We use the shape-aware data augmentation method proposed by SE-SSD \cite{zheng2021se}. This module simplifies the handling of partial occlusions, sparsity and different shapes of objects in the same class. Some traditional augmentation methods are also applied before the shape-aware augmentation, e.g. rotation, flipping, and scaling. For the generation of semi-synthetic data, several data augmentation techniques are also implemented to increase the variance of the point clouds. Therefore, in every second frame, 0-20\% of all points are dropped out and Gaussian noise with $\sigma=0.2$ is added to 20-40\% of all points. These techniques increase the variance of point clouds and provide more robust and diverse data. Data augmentation plays an important role for domain adaptation methods, as it is used by several methods \cite{ST3D}, \cite{completeandlabel},\cite{xMUDA}. It is noticeable, that the point density of the target domain is more important than the point density of the source domain \cite{ST3D}. Subsequently, cropping and drop out of points, respectively point cloud upsampling is an important step to adjust the number of points of the source set to the target set. Statistics for both, the source and the target domain dataset, are calculated with respect to the number of points of the total point cloud and, if annotated data is available for the target domain data, the average number of points per object. Then, the data augmentation techniques drop out and upsampling are used to match the source domain dataset to the target domain dataset. To better illustrate this effect, a domain adaptation is applied in \cref{evaluation} from the synthetic A9 dataset (source domain) to the A11 and D16 dataset of the \textit{Regensburg Next} project (target domain). Note that the number of points for the target domain set is reduced by a factor of 2.72 compared to the source domain set, whereas the average number of points for the \textit{Car} class is increased by a factor of 1.83 compared to the source dataset. Due to the different LiDARs used for both smart intersections, there is more overlap between the four permanently installed Blickfeld Cube 1 LiDARs in the \textit{Regensburg Next} project. This sensor-to-sensor domain adaptation is considered in more detail in \cref{evaluation}.

\textbf{Self-Ensembling Training Framework.}
We introduce the self-ensembling training framework \cite{zheng2021se} to do a post training: We first train the model without self-ensembling, and then we take the pre-trained model as a teacher model to train the student model that has the same network structure. Predictions of the teacher model are used as soft supervision. Combined with the hard supervision from the ground truth, we can provide more information to the student model. The student and teacher model are initialized with the same pre-trained parameters. The generated soft targets are obtained by the teacher SSD. They include a global transformation that performs translation, flipping and scaling as data augmentation techniques. After a global transformation, shape-aware data augmentation is performed on input points with the corresponding ground truth annotations (hard targets). The augmented input is fed into the student SSD together with the consistency loss that is obtained by the teacher and student predictions in order to align the student predictions with the provided soft targets.

Furthermore, the student is supervised with the orientation-aware distance-IoU loss to better exploit hard targets for regression bounding boxes. The overall loss for training the student model consists of:
\begin{equation}
    \mathcal{L}_{student} =\mathcal{L}_{cls}^s + \omega_1\mathcal{L}_{OD-IoU}^s +\omega_2\mathcal{L}_{dir}^s + \mu_t\mathcal{L}_{consist},
    \label{eq:se_training_loss_}
\end{equation}
where $\mathcal{L}_{cls}^s$ is the focal loss \cite{lin2017focal} for box classification, $\mathcal{L}_{OD-IoU}^s$ is the OD-IoU loss for bounding box regression, $\mathcal{L}_{dir}^s $ is the cross-entropy loss for direction classification, $\mathcal{L}_{consist}$ is the consistency loss, that is a sum of the bounding box loss and the classification loss, $\omega_1$, $\omega_2$, $\lambda$ and $\mu_t$ are weights of losses.
During post-training, the parameters of the teacher model are updated with the ones of the student model using the exponential moving average (EMA) strategy. 

\section{Evaluation}\label{evaluation}
To prove the effectiveness of our proposed modules in \textit{DASE-ProPillars}, we evaluate the model on the A9 dataset which is a novel roadside dataset in the autonomous driving domain with labeled roadside sensor data. Furthermore, we finetune and evaluate the model on two further roadside LiDAR datasets, the synthetic A11 and D16 datasets from the \textit{Regensburg Next} project. The training and evaluation was performed on 3 x NVIDIA GeForce RTX 3090 GPUs.

\subsection{A9 dataset}
We use the \textit{DASE-ProPillars} model to train on the training set of the A9 dataset for 80 epochs with a batch size of 8, and evaluate on the test set. As optimizer, Adam is used with a weight decay of 0.01. The learning rate is of type "one\_cycle" with a maximum of 0.003. We report our result using 3D and BEV under IoU threshold 0.5 and 0.25 like in \cite{sepropillars}. The confidence score is limited with threshold 0.1, but we again decrease the NMS IoU threshold to 0.2, exactly the same setting as the convention in nuScenes \cite{caesar2020nuscenes}. The LiDAR frames of the second batch of the A9 dataset are labeled on the intersection. We test the case of using 0.1 as the NMS threshold. The inference time is 22.0 ms that still provides real-time object detection results. The result is shown in \cref{tab:evaluation_a9_dataset}.

\begin{table}[htbp]
\centering
\scalebox{0.95}{
\begin{tabular}{|l|cc|cc|}
\hline
Metric     & \multicolumn{2}{c|}{3D mAP}        & \multicolumn{2}{c|}{BEV mAP}        \\ \hline\hline
IoU threshold & \multicolumn{1}{c|}{0.5} & 0.25 & \multicolumn{1}{c|}{0.5} & 0.25 \\ \hline \hline
SE-ProPillars & \multicolumn{1}{c|}{30.13}   & 50.09   & \multicolumn{1}{c|}{40.21}   & 51.53 \\ \hline
DASE-ProPillars (Ours) & \multicolumn{1}{c|}{\textbf{54.38}} & \textbf{80.65} & \multicolumn{1}{c|}{\textbf{55.10}} & \textbf{83.38} \\ \hline
\end{tabular}
}
\caption{3D object detection results of \textit{DASE-ProPillars} on the A9 test set. We report the 3D and BEV mAP of Car under 0.5 and 0.25 IoU threshold, with 40 recall positions.}
\label{tab:evaluation_a9_dataset}
\end{table}

\subsection{A11 and D16 dataset}
The results of the \textit{DASE-ProPillars} model trained on the semi-synthetic \textit{proSynthSemi} dataset are used to improve the results on similar roadside datasets, the A11 and D16 datasets from the \textit{Regensburg Next} project \cite{torunskyPilotprojektMitVorbildcharakter}, which consist of 2,581 and 3,315 synthetic frames respectively. The simulation environment \textit{Gazebo} was used to randomly place objects in a predefined region of an intersection to generate annotated data. The LiDARs in the real A9 dataset and the semi-synthetic A9 dataset are mounted on traffic sign gantry bridges, whereas the LiDAR sensors in the synthetic A11 dataset are mounted on roadside rooftops in a height of approximately 7 m above the ground plane. We use our \textit{DASE-ProPillars} model to train on the training set of the synthetic A9 dataset for 80 epochs and finetune on the A11 dataset for 40 epochs, with a decrease of the learning rate of 50\%. For comparison reasons we also trained our \textit{DASE-ProPillars} model directly on the A11 dataset for 40 epochs using the same hyperparameters as above. The NMS IoU threshold is decreased to 0.2 for both models. As mentioned above, the average number of points per object is increased by a factor of 1.83 compared to the synthetic A9 dataset. Therefore, point upsampling was used, to match the number of average points per objects for both domains. Since the average number of points of the total point cloud in the target set is significantly lower (factor 2.72) compared to the source set, only points of the objects are upsampled. In order to reduce the difference in terms of total number of points of the respective point clouds simultaneously, drop out is used to reduce the number of total points of the source set. \Cref{tab:evaluation_regensburg_next_dataset} shows the impact of several data augmentation techniques of a \textit{DASE-ProPillars} model trained only on the source set and evaluated directly on the target set without the use of any finetuning. Upsampling object points improved the 3D object detection, whereas the subsequent drop out had led to a slight deterioration, contrary to expectations. However, the drop out rate was set to 0.5, meaning that 50\% of all points were dropped. Using a smaller drop out rate of 0.25, the results compared to the higher rate increased. \Cref{tab:evaluation_rg_finetuning} shows the results for the models trained only on the A11 dataset of the \textit{Regensburg Next} project compared with the model which was trained on the synthetic A9 dataset and finetuned on the target domain set.

\begin{table}[htbp]
\centering
\begin{tabular}{|l|c|c|}
\hline
Metric & 3D mAP &  BEV mAP \\ \hline\hline
upsampling + drop out (0.5) & 49.64 & 61.14 \\ \hline
drop out (0.25)  & 60.92 & 69.05 \\ \hline
original & 61.72 & 72.83 \\ \hline
upsampling & \textbf{63.11} & \textbf{76.65} \\ \hline
\end{tabular}
\caption{3D object detection results of \textit{DASE-ProPillars} on the test set of the A11 dataset. We report the 3D and BEV mAP of the \textit{Car} class under an IoU threshold of 0.25, with 40 recall positions including several data augmentation techniques.}
\label{tab:evaluation_regensburg_next_dataset}
\end{table}
\begin{table}[htbp]
\centering
\scalebox{0.85}{
\begin{tabular}{|l|cc|cc|}
\hline
Metric     & \multicolumn{2}{c|}{3D mAP}        & \multicolumn{2}{c|}{BEV mAP}        \\\cline{2-5}
IoU threshold & \multicolumn{1}{c|}{0.5} & 0.25 & \multicolumn{1}{c|}{0.5} & 0.25 \\ \hline \hline
DASE-ProPillars  & \multicolumn{1}{c|}{23.26}   & 89.87   & \multicolumn{1}{c|}{27.61}   & 93.42 \\ \hline
DASE-ProPillars + fine-tuning & \multicolumn{1}{c|}{\textbf{33.73}} & \textbf{93.49} & \multicolumn{1}{c|}{\textbf{38.81}} & \textbf{94.79} \\ \hline

\end{tabular}
}
\caption{3D object detection results of \textit{DASE-ProPillars} on the test set of the A11 dataset. We report the 3D and BEV mAP of Car under 0.5 and 0.25 IoU threshold, with 40 recall positions.}
\label{tab:evaluation_rg_finetuning}
\end{table}

\section{Conclusion}
In this work we presented our \textit{DASE-ProPillars} 3D object detector that is an improved version of the \textit{PointPillars} model. We show generalization ability of our model and make it more robust via domain adaptation. We replace the detection head with a more lightweight multi-task head. We add two training techniques to our baseline: the shape-aware data augmentation module and the self-ensembling training architecture. Sufficient data collection is key to train a model and achieve a good accuracy. We do several sets of experiments for each module to prove its accuracy and runtime performance.\\
To sum up, the \textit{DASE-ProPillars} 3D object detector is a significant contribution within the area of LiDAR-based 3D perception to support self-driving vehicles and improve road traffic safety.



\section*{APPENDIX}


\section{Detailed Module Architecture}

\subsection{Attentive Hierarchical Middle Layers}
\begin{figure*}[t]
    \centering
    \includegraphics[width=1.0\linewidth,trim=0 10cm 0 0]{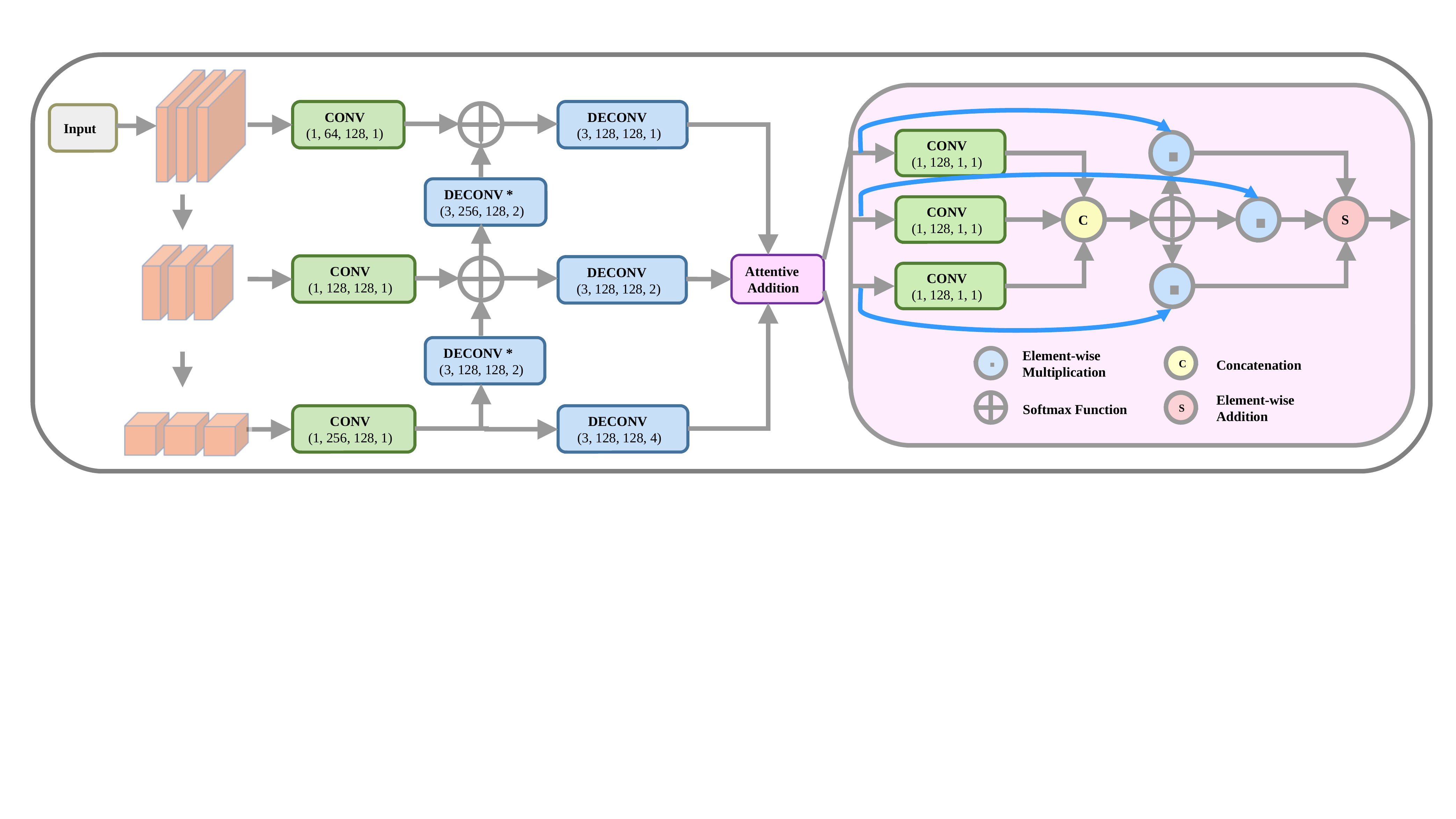}
    \caption{Left: Structure of the attentive hierarchical middle layers. Right: Structure of the attentive addition operation.}
    \label{fig:attentive_hierarchical_middle_layers}
\end{figure*}

We exchange the default backbone of \textit{PointPillars} with an \textit{Attentive Hierarchical Backbone} to perform 2D convolution on the pseudo image from the pillar feature net. Figure \ref{fig:attentive_hierarchical_middle_layers} depicts the structure of the attentive hierarchical middle layers. 

In the first stage, the spatial resolution of the pseudo image is gradually downsampled by three groups of convolutions. Each group contains three convolutional layers, where the first one has a stride of two for downsampling, and the two subsequent layers act only for feature extraction. After downsampling, deconvolution operations are applied to recover the spatial resolution. Deconvolutional layers (marked with an asterix) recover the size of feature maps with stride 2 and element-wise add them to upper branches. The remaining three deconvolutional layers make all three branches have the same size (half of the original feature map). Then the final three feature maps are combined by an attentive addition to fuse both, spatial and semantic features. The attentive addition uses the plain attention mechanism. All three feature maps are being passed through a convolutional operation and are channel-wise concatenated as attention scores. The softmax function generates the attention distribution and feature maps are multiplied with the corresponding distribution weight. The element-wise addition in the end gives the final attention output, a ($C \times H/2 \times W/2$) feature map.

\subsection{Self-Ensembling Training Framework}
\begin{figure*}[t]
    \centering
      \includegraphics[width=0.9\linewidth]{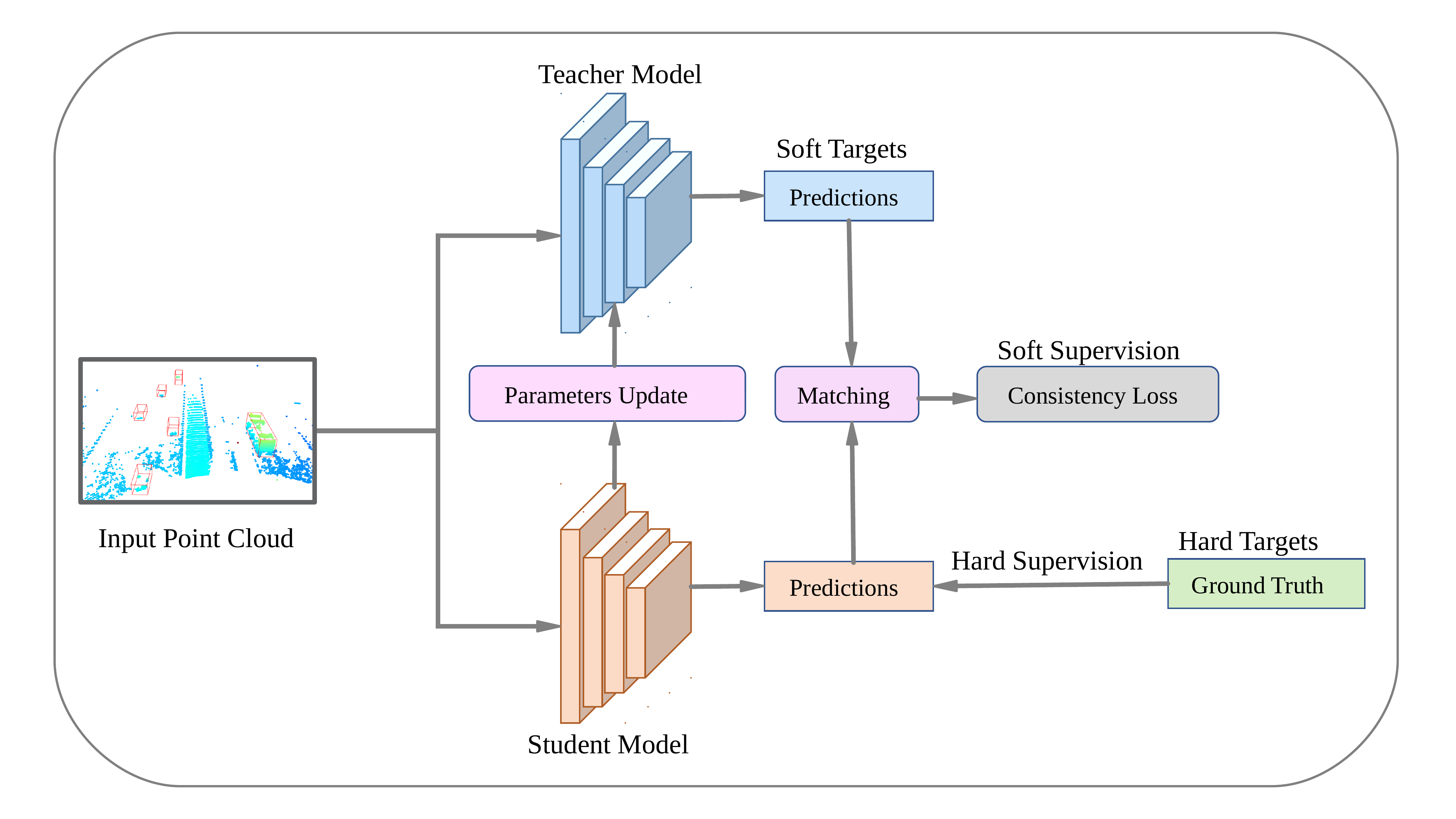}
    \caption{Self-ensembling training architecture.}
    \label{fig:teacher_student_architecture}
\end{figure*}

In addition, we introduce the self-ensembling training framework \cite{zheng2021se} to do a post training: We first train the model shown in Fig. \ref{fig:teacher_student_architecture} but without the self-ensembling module, and then we take the pre-trained model as a teacher model to train the student model that has the same network structure. 

\section{Semi-Synthetic Data Generation}

In the CARLA simulator, the frequency of a simulated LiDAR has to be set to 20 Hz due to CARLA time synchronization problems. Using 10 Hz, the generated point cloud is cut off at half for each frame. The data acquisition with 20 Hz solves this problem to generate a complete point cloud. However, not as expected 20 frames per second, but only 10 frames per second are captured, what coincides with a data acquisition at 10 Hz. With an initial horizontal resolution of $2,048$, the number of points have to be doubled to $2,621,480$. To reduce the \textit{sim-to-real} gap, noise with a standard deviation of $0.1$ and a general drop-off of $10$ \% is included to the generated point cloud directly in the acquisition process. The characteristics of the simulated LiDAR are depicted in \cref{tab:charactersistics_simulated_lidar}.

Using a semi-synthetic data generation pipeline, a synthetically generated point cloud containing automatically annotated data can be included into a point cloud which is captured in real world. The main concept of the semi-synthetic data generation is to insert synthetically generated points of an object, for which the annotation is available, into a real point cloud. In this context, one has to ensure, that the real world and the simulated LiDAR cover the same map region. Furthermore, the proportions of the environment must match between real world and simulation. Using \textit{OpenDRIVE}, roads can be well integrated into the simulation and GPS measured points can be transferred to the CARLA simulation, which guarantees, that the LiDAR positions in real world and simulation are identical. If these conditions for the simulation are met, the advantages of synthetic and real data can be combined and only synthetic generated point clouds and a minimal real point cloud, which does not contain any objects, is required. To add more variance to the data, multiple real point clouds without any objects can be included. The semi-synthetic data generation can be split up into the following steps:\\

\begin{enumerate}
    \item Extraction of synthetic object points from synthetic data
    \item Identifying the ground plane via RANSAC
    \item Creation of a height profile
    \item Removal of ground points
    \item Including of synthetic objects point into real point cloud
    \item Application of data augmentation
    \item Adjustment of annotations
\end{enumerate}   
\begin{table}[b]
\centering
\begin{tabular}{l|c}
\hline
Actor attribute     & Value          \\ \hline
Channels &  64    \\ \hline 
Range & 120 m \\ \hline
Points per Second  & 2,621,480 \\ \hline
Rotation Rate   & 20 Hz  \\ \hline
Vertical FOV  & $45^{\circ}$ ($\pm 22.5^{\circ}$) \\ \hline
Horizontal FOV & $360^{\circ}$ \\ \hline
Noise & 0.1 \\ \hline
Dropoff general rate & 0.1 \\ \hline
\end{tabular}
\caption{Characteristics of the simulated OS1-64 (gen. 2).\label{tab:charactersistics_simulated_lidar}}
\end{table}

\begin{figure*}[h]
\centering
\minipage{0.33\textwidth}
  \includegraphics[width=\linewidth]{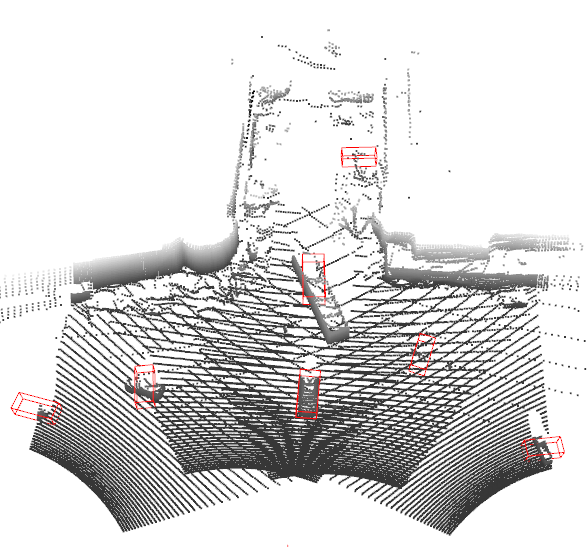}
\endminipage
\minipage{0.33\textwidth}
  \includegraphics[width=\linewidth]{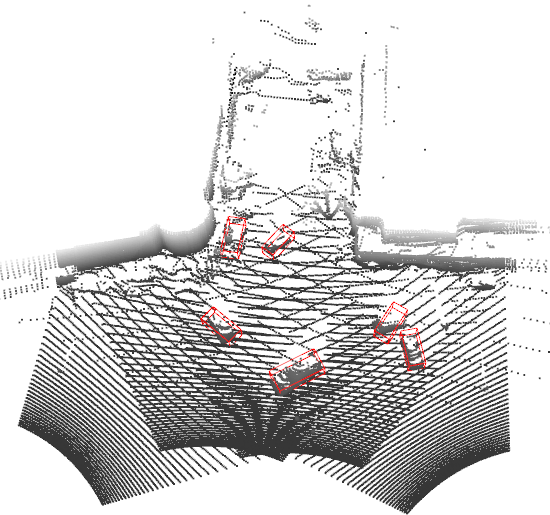}
\endminipage
\minipage{0.33\textwidth}%
  \includegraphics[width=\linewidth]{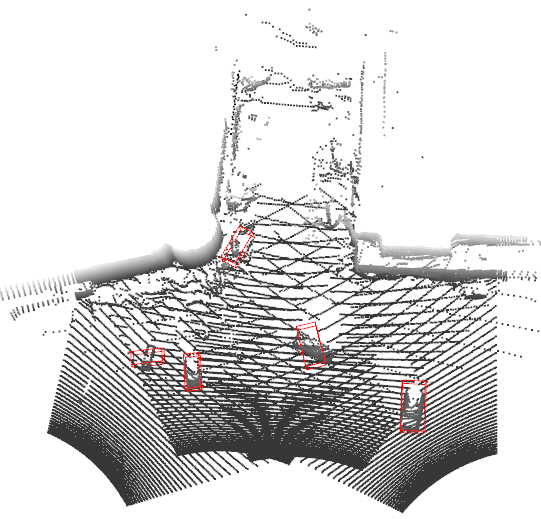}
\endminipage
\hfill
\minipage{0.33\textwidth}
  \includegraphics[width=\linewidth]{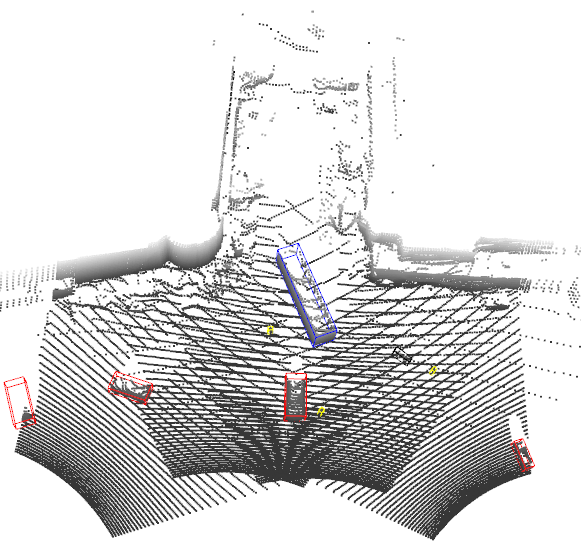}
  \caption*{(a) MonoDet3D}
\endminipage
\minipage{0.33\textwidth}
  \includegraphics[width=\linewidth]{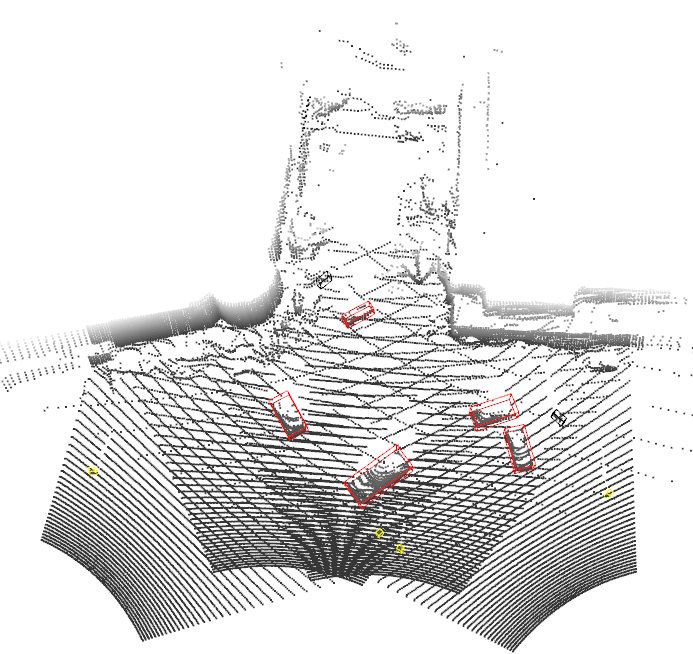}
  \caption*{(b) PointPillars}
\endminipage
\minipage{0.33\textwidth}%
  \includegraphics[width=\linewidth]{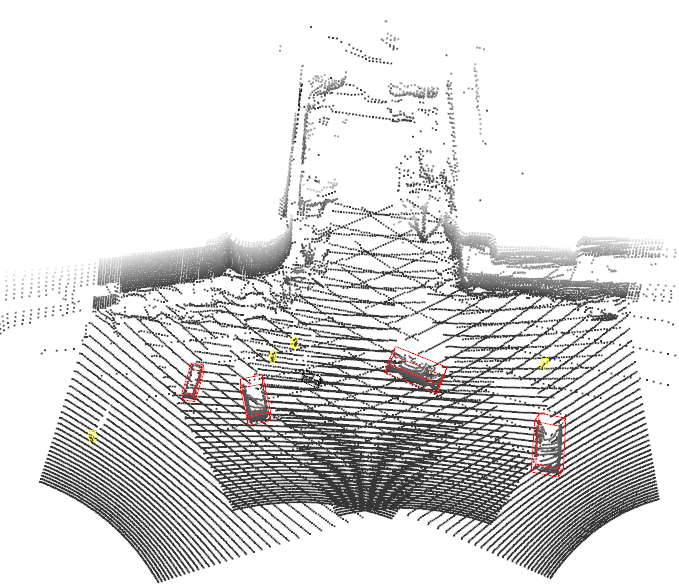}
  \caption*{(c) InfraDet3D}
\endminipage
\caption{Top row: Example frames of the target domain (A11\_dataset) including predictions produced by the DASE-ProPillars object detector using cross-sensor domain adaptation. Bottom row: Corresponding ground truth annotations (car: red; bus: blue; pedestrian: yellow, motorbike: black).}%
\label{fig:eval_examples_da_cross_sensor}%
\end{figure*}

\section{Additional Ablation Studies}
Table \ref{tab:eval_da_syn_to_real} lists the 3D, BEV and AOS mAP of the car class on the $s110\_v01$ dataset under 0.25 IoU threshold with 40 recall positions for both, the initial baseline without any domain adaptation and the step-by-step enabling of all domain adaptation modules. It can be seen, that including all domain adaptation processes outperforms the initial baseline approach in all metrics significantly, having the same run-time of 20 ms. The transformation of the synthetic to semi-synthetic data increases the BEV and AOS metric by 51.19 and 35.02 respectively.

\begin{table}[h]
\centering
\begin{tabular}{l|c|c|c}
\hline
$mAP_{40}@0.25$ & BEV & 3D  & AOS   \\ \hline \hline
proSynthLiDAR  & 3.03  & 1.84 & 1.50  \\ \hline 
baseline (proSemiSynthLiDAR)  & 54.22  & 7.84 & 36.52  \\ \hline 
baseline + VA & 63.65  & 12.40 & 41.15  \\ \hline
baseline + VA + FT & 65.12  & 51.60 & 44.32  \\ \hline
baseline + VA + FT + BN & \textbf{67.69} & \textbf{61.84} & \textbf{46.09} \\ \hline
\end{tabular}
\caption{Ablation study synthetic-to-real domain adaptation. The modules of the synthetic-to-real domain adaptation are enabled step by step. First, the impact of the semi-synthetic data is highlighted. Afterwards, the voxel-alignment (VA), fine-tuning (FT) and the bounding box normalization (BN) modules are investigated. We report the BEV, 3D and AOS mAP for the class car under 0.25 IoU threshold, with 40 recall positions.}
\label{tab:eval_da_syn_to_real}
\end{table}

\section{Qualitative Evaluation Results}
Figure \ref{fig:eval_examples_da_cross_sensor} provides several examples of the target domain set including the predictions of the \textit{DASE-ProPillars} model and the corresponding ground truths. In the first frame, the bus and the motorbike are detected wrongly as a car and also a part of the environment is predicted and therefore denotes one false positive. The remaining two frames also include one false positive as a bush on the roadside is detected as a car. However, in all frames, all annotated cars are detected correctly and also the rotation fits accurately for most objects. The visual representation of the predictions emphasizes the results of the proposed \textit{DASE-ProPillars} object detector.

\addtolength{\textheight}{-18.5cm}   

\begin{thebibliography}{10}
\providecommand{\url}[1]{#1}
\csname url@samestyle\endcsname
\providecommand{\newblock}{\relax}
\providecommand{\bibinfo}[2]{#2}
\providecommand{\BIBentrySTDinterwordspacing}{\spaceskip=0pt\relax}
\providecommand{\BIBentryALTinterwordstretchfactor}{4}
\providecommand{\BIBentryALTinterwordspacing}{\spaceskip=\fontdimen2\font plus
\BIBentryALTinterwordstretchfactor\fontdimen3\font minus
  \fontdimen4\font\relax}
\providecommand{\BIBforeignlanguage}[2]{{%
\expandafter\ifx\csname l@#1\endcsname\relax
\typeout{** WARNING: IEEEtran.bst: No hyphenation pattern has been}%
\typeout{** loaded for the language `#1'. Using the pattern for}%
\typeout{** the default language instead.}%
\else
\language=\csname l@#1\endcsname
\fi
#2}}
\providecommand{\BIBdecl}{\relax}
\BIBdecl

\bibitem{cress2022a9}
C.~Cre{\ss}, W.~Zimmer, L.~Strand, M.~Fortkord, S.~Dai, V.~Lakshminarasimhan,
  and A.~Knoll, ``A9-dataset: Multi-sensor infrastructure-based dataset for
  mobility research,'' \emph{arXiv preprint arXiv}, 2022.

\bibitem{lang2019pointpillars}
A.~H. Lang, S.~Vora, H.~Caesar, L.~Zhou, J.~Yang, and O.~Beijbom,
  ``Pointpillars: Fast encoders for object detection from point clouds,'' in
  \emph{Proceedings of the IEEE/CVF Conference on Computer Vision and Pattern
  Recognition}, 2019, pp. 12\,697--12\,705.

\bibitem{zheng2021se}
W.~Zheng, W.~Tang, L.~Jiang, and C.-W. Fu, ``Se-ssd: Self-ensembling
  single-stage object detector from point cloud,'' in \emph{Proceedings of the
  IEEE/CVF Conference on Computer Vision and Pattern Recognition}, 2021, pp.
  14\,494--14\,503.

\bibitem{liu2020tanet}
Z.~Liu, X.~Zhao, T.~Huang, R.~Hu, Y.~Zhou, and X.~Bai, ``Tanet: Robust 3d
  object detection from point clouds with triple attention,'' in
  \emph{Proceedings of the AAAI Conference on Artificial Intelligence},
  vol.~34, no.~07, 2020, pp. 11\,677--11\,684.

\bibitem{krammer2022providentia}
A.~Kr{\"a}mmer, C.~Sch{\"o}ller, D.~Gulati, V.~Lakshminarasimhan, F.~Kurz,
  D.~Rosenbaum, C.~Lenz, and A.~Knoll, ``Providentia-a large-scale sensor
  system for the assistance of autonomous vehicles and its evaluation,''
  \emph{Journal of Field Robotics}, 2022.

\bibitem{lakshminarasimhan2020c}
V.~Lakshminarasimhan and A.~Knoll, ``C-v2x resource deployment architecture
  based on moving network convoys,'' in \emph{2020 IEEE 91st vehicular
  technology conference (VTC2020-Spring)}.\hskip 1em plus 0.5em minus
  0.4em\relax IEEE, 2020, pp. 1--6.

\bibitem{cress2021intelligent}
C.~Cre{\ss} and A.~C. Knoll, ``Intelligent transportation systems with the use
  of external infrastructure: A literature survey,'' \emph{arXiv preprint
  arXiv:2112.05615}, 2021.

\bibitem{torunskyPilotprojektMitVorbildcharakter}
V.~R. Torunsky, ``{Pilotprojekt mit Vorbildcharakter},'' p.~1.

\bibitem{dosovitskiy2017carla}
A.~Dosovitskiy, G.~Ros, F.~Codevilla, A.~Lopez, and V.~Koltun, ``Carla: An open
  urban driving simulator,'' in \emph{Conference on robot learning}.\hskip 1em
  plus 0.5em minus 0.4em\relax PMLR, 2017, pp. 1--16.

\bibitem{OpenLABELStandard}
``Asam e.v. openlabel v1.0.0 standardization project.''
  https://www.asam.net/project-detail/asam-openlabel-v100/.

\bibitem{sepropillars}
J.~Wu, W.~Zimmer, and A.~Knoll, ``Real-time lidar-based 3d object detection on
  the providentia++ test stretch using a single-stage architecture,'' Master's
  thesis, Technische Universität München, 2021, unpublished thesis.

\bibitem{walter2022a}
W.~Zimmer, E.~Ercelik, X.~Zhou, X.~Jair Diaz~Ortiz, and A.~Knoll, ``A survey of
  robust 3d object detection methods in point clouds,'' \emph{arXiv preprint
  arXiv:submit/4161670}, 2022.

\bibitem{pointrcnn}
S.~Shi, X.~Wang, and H.~Li, ``Pointrcnn: 3d object proposal generation and
  detection from point cloud,'' in \emph{Proceedings of the IEEE/CVF conference
  on computer vision and pattern recognition}, 2019, pp. 770--779.

\bibitem{pointnet++}
C.~R. Qi, L.~Yi, H.~Su, and L.~J. Guibas, ``Pointnet++ deep hierarchical
  feature learning on point sets in a metric space,'' in \emph{Proceedings of
  the 31st International Conference on Neural Information Processing Systems},
  2017, pp. 5105--5114.

\bibitem{3dssd}
Z.~Yang, Y.~Sun, S.~Liu, and J.~Jia, ``3dssd: Point-based 3d single stage
  object detector,'' in \emph{Proceedings of the IEEE/CVF conference on
  computer vision and pattern recognition}, 2020, pp. 11\,040--11\,048.

\bibitem{second}
Y.~Yan, Y.~Mao, and B.~Li, ``Second: Sparsely embedded convolutional
  detection,'' \emph{Sensors}, vol.~18, no.~10, p. 3337, 2018.

\bibitem{subsparseCNN}
B.~Graham, M.~Engelcke, and L.~Van Der~Maaten, ``3d semantic segmentation with
  submanifold sparse convolutional networks,'' in \emph{Proceedings of the IEEE
  conference on computer vision and pattern recognition}, 2018, pp. 9224--9232.

\bibitem{sassd}
C.~He, H.~Zeng, J.~Huang, X.-S. Hua, and L.~Zhang, ``Structure aware
  single-stage 3d object detection from point cloud,'' in \emph{Proceedings of
  the IEEE/CVF Conference on Computer Vision and Pattern Recognition}, 2020,
  pp. 11\,873--11\,882.

\bibitem{zheng2020cia}
W.~Zheng, W.~Tang, S.~Chen, L.~Jiang, and C.-W. Fu, ``Cia-ssd: Confident
  iou-aware single-stage object detector from point cloud,'' \emph{arXiv
  preprint arXiv:2012.03015}, 2020.

\bibitem{mao2021pyramid}
J.~Mao, M.~Niu, H.~Bai, X.~Liang, H.~Xu, and C.~Xu, ``Pyramid r-cnn: Towards
  better performance and adaptability for 3d object detection,'' in
  \emph{Proceedings of the IEEE/CVF International Conference on Computer
  Vision}, 2021, pp. 2723--2732.

\bibitem{sun2020scalability}
P.~Sun, H.~Kretzschmar, X.~Dotiwalla, A.~Chouard, V.~Patnaik, P.~Tsui, J.~Guo,
  Y.~Zhou, Y.~Chai, B.~Caine \emph{et~al.}, ``Scalability in perception for
  autonomous driving: Waymo open dataset,'' in \emph{Proceedings of the
  IEEE/CVF conference on computer vision and pattern recognition}, 2020, pp.
  2446--2454.

\bibitem{VoTr}
J.~Mao, Y.~Xue, M.~Niu, H.~Bai, J.~Feng, X.~Liang, H.~Xu, and C.~Xu, ``Voxel
  transformer for 3d object detection,'' 2021.

\bibitem{BtcDet}
\BIBentryALTinterwordspacing
Q.~Xu, Y.~Zhong, and U.~Neumann, ``Behind the curtain: Learning occluded shapes
  for 3d object detection,'' \emph{CoRR}, vol. abs/2112.02205, 2021. [Online].
  Available: \url{https://arxiv.org/abs/2112.02205}
\BIBentrySTDinterwordspacing

\bibitem{TransferLearningFromSynthToReal}
\BIBentryALTinterwordspacing
A.~Xiao, J.~Huang, D.~Guan, F.~Zhan, and S.~Lu, ``Synlidar: Learning from
  synthetic lidar sequential point cloud for semantic segmentation,''
  \emph{CoRR}, vol. abs/2107.05399, 2021. [Online]. Available:
  \url{https://arxiv.org/abs/2107.05399}
\BIBentrySTDinterwordspacing

\bibitem{surveyDA}
\BIBentryALTinterwordspacing
L.~T. Triess, M.~Dreissig, C.~B. Rist, and J.~M. Z{\"{o}}llner, ``A survey on
  deep domain adaptation for lidar perception,'' \emph{CoRR}, vol.
  abs/2106.02377, 2021. [Online]. Available:
  \url{https://arxiv.org/abs/2106.02377}
\BIBentrySTDinterwordspacing

\bibitem{domainandmodalitygaps2021}
\BIBentryALTinterwordspacing
D.~Jia, A.~Hermans, and B.~Leibe, ``Domain and modality gaps for lidar-based
  person detection on mobile robots,'' \emph{CoRR}, vol. abs/2106.11239, 2021.
  [Online]. Available: \url{https://arxiv.org/abs/2106.11239}
\BIBentrySTDinterwordspacing

\bibitem{traininger}
\BIBentryALTinterwordspacing
Y.~Wang, X.~Chen, Y.~You, L.~E. Li, B.~Hariharan, M.~E. Campbell, K.~Q.
  Weinberger, and W.~Chao, ``Train in germany, test in the {USA:} making 3d
  object detectors generalize,'' \emph{CoRR}, vol. abs/2005.08139, 2020.
  [Online]. Available: \url{https://arxiv.org/abs/2005.08139}
\BIBentrySTDinterwordspacing

\bibitem{ST3D}
\BIBentryALTinterwordspacing
J.~Yang, S.~Shi, Z.~Wang, H.~Li, and X.~Qi, ``{ST3D:} self-training for
  unsupervised domain adaptation on 3d object detection,'' \emph{CoRR}, vol.
  abs/2103.05346, 2021. [Online]. Available:
  \url{https://arxiv.org/abs/2103.05346}
\BIBentrySTDinterwordspacing

\bibitem{ransac}
K.~G. Derpanis, ``Overview of the ransac algorithm,'' \emph{Image Rochester
  NY}, vol.~4, no.~1, pp. 2--3, 2010.

\bibitem{eldar1997farthest}
Y.~Eldar, M.~Lindenbaum, M.~Porat, and Y.~Y. Zeevi, ``The farthest point
  strategy for progressive image sampling,'' \emph{IEEE Transactions on Image
  Processing}, vol.~6, no.~9, pp. 1305--1315, 1997.

\bibitem{hu2018squeeze}
J.~Hu, L.~Shen, and G.~Sun, ``Squeeze-and-excitation networks,'' in
  \emph{Proceedings of the IEEE conference on computer vision and pattern
  recognition}, 2018, pp. 7132--7141.

\bibitem{he2016deep}
K.~He, X.~Zhang, S.~Ren, and J.~Sun, ``Deep residual learning for image
  recognition,'' in \emph{Proceedings of the IEEE conference on computer vision
  and pattern recognition}, 2016, pp. 770--778.

\bibitem{qi2017pointnet}
C.~R. Qi, H.~Su, K.~Mo, and L.~J. Guibas, ``Pointnet: Deep learning on point
  sets for 3d classification and segmentation,'' in \emph{Proceedings of the
  IEEE conference on computer vision and pattern recognition}, 2017, pp.
  652--660.

\bibitem{ioffe2015batch}
S.~Ioffe and C.~Szegedy, ``Batch normalization: Accelerating deep network
  training by reducing internal covariate shift,'' in \emph{International
  conference on machine learning}.\hskip 1em plus 0.5em minus 0.4em\relax PMLR,
  2015, pp. 448--456.

\bibitem{nair2010rectified}
V.~Nair and G.~E. Hinton, ``Rectified linear units improve restricted boltzmann
  machines,'' in \emph{Icml}, 2010.

\bibitem{yan2018second}
Y.~Yan, Y.~Mao, and B.~Li, ``Second: Sparsely embedded convolutional
  detection,'' \emph{Sensors}, vol.~18, no.~10, p. 3337, 2018.

\bibitem{completeandlabel}
\BIBentryALTinterwordspacing
L.~Yi, B.~Gong, and T.~A. Funkhouser, ``Complete {\&} label: {A} domain
  adaptation approach to semantic segmentation of lidar point clouds,''
  \emph{CoRR}, vol. abs/2007.08488, 2020. [Online]. Available:
  \url{https://arxiv.org/abs/2007.08488}
\BIBentrySTDinterwordspacing

\bibitem{xMUDA}
\BIBentryALTinterwordspacing
M.~Jaritz, T.~Vu, R.~de~Charette, {\'{E}}.~Wirbel, and P.~P{\'{e}}rez, ``xmuda:
  Cross-modal unsupervised domain adaptation for 3d semantic segmentation,''
  \emph{CoRR}, vol. abs/1911.12676, 2019. [Online]. Available:
  \url{http://arxiv.org/abs/1911.12676}
\BIBentrySTDinterwordspacing

\bibitem{lin2017focal}
T.-Y. Lin, P.~Goyal, R.~Girshick, K.~He, and P.~Doll{\'a}r, ``Focal loss for
  dense object detection,'' in \emph{Proceedings of the IEEE international
  conference on computer vision}, 2017, pp. 2980--2988.

\bibitem{caesar2020nuscenes}
H.~Caesar, V.~Bankiti, A.~H. Lang, S.~Vora, V.~E. Liong, Q.~Xu, A.~Krishnan,
  Y.~Pan, G.~Baldan, and O.~Beijbom, ``nuscenes: A multimodal dataset for
  autonomous driving,'' in \emph{Proceedings of the IEEE/CVF conference on
  computer vision and pattern recognition}, 2020, pp. 11\,621--11\,631.

\end{thebibliography}


\end{document}